\pdfoutput=1

\documentclass[11pt]{article}

\usepackage[]{acl}

\usepackage{times}
\usepackage{latexsym}
\usepackage{float} 
\usepackage[T1]{fontenc}

\usepackage[utf8]{inputenc}

\usepackage{microtype}

\usepackage{booktabs}
\usepackage{tabularx}
\usepackage{graphicx}
\newcommand{\ee}{{\sc EditEval}}
\usepackage{enumitem}
\usepackage{amsmath}
\usepackage{caption}
\usepackage{subcaption}

\title{\ee: An Instruction-Based Benchmark for Text Improvements}

\author{Jane Dwivedi-Yu$^{\diamondsuit}$ \quad Timo Schick$^{\diamondsuit}$ \quad Zhengbao Jiang$^{\diamondsuit,\heartsuit}$ \\[4pt] { \bf Maria Lomeli$^{\diamondsuit}$ \quad Patrick Lewis$^{\diamondsuit}$ \quad \bf Gautier Izacard$^{\diamondsuit,\clubsuit}$ \quad } \\[4pt] { \bf Edouard Grave$^{\diamondsuit}$\quad Sebastian Riedel$^{\diamondsuit,\spadesuit}$ \quad Fabio Petroni$^{\diamondsuit}$} \\[8pt]
$^{\diamondsuit}$ Meta AI Research,
$^{\heartsuit}$ Carnegie Mellon University,\\
$^{\clubsuit}$ Inria \& ENS, PSL University,
$^{\spadesuit}$ University College London \\[4pt]
{\tt \{janeyu,schick,zhengbao,marialomeli,plewis,gizacard}\\
{\tt egrave,sriedel,fabiopetroni\}@meta.com}
}
\begin{document}
\maketitle
\begin{abstract}
Evaluation of text generation to date has primarily focused on content created sequentially, rather than improvements on a piece of text. Writing, however, is naturally an iterative and incremental process that requires expertise in different modular skills such as fixing outdated information or making the style more consistent. Even so, comprehensive evaluation of a model's capacity to perform these skills and the ability to edit remains sparse. This work presents \ee: An instruction-based, benchmark and evaluation suite that leverages high-quality existing and new datasets for automatic evaluation of editing capabilities such as making text more cohesive and paraphrasing. We evaluate several pre-trained models, which shows that InstructGPT and PEER perform the best, but that most baselines fall below the supervised SOTA, particularly when neutralizing and updating information. Our analysis also shows that commonly used metrics for editing tasks do not always correlate well, and that optimization for prompts with the highest performance does not necessarily entail the strongest robustness to different models. Through the release of this benchmark,\footnote{Code and data available at \url{https://github.com/facebookresearch/EditEval}} and a publicly available leaderboard challenge,\footnote{\url{https://eval.ai/web/challenges/challenge-page/1866/overview}} we hope to unlock future research in developing models capable of iterative and more controllable editing.
\end{abstract}


\section{Introduction}

Large pre-trained language models have shown impressive text generation capabilities for a wide variety of tasks such as question answering, textual entailment, and summarization~\cite{devlin2019bert, radford2019language, raffel2020exploring, brown2020language, zhang2022opt, chowdhery2022palm}. However, to date, most work employing language models has focused on generating immutable text in a single pass. This is in stark contrast to the way in which humans develop articles of text, which is naturally an iterative process of small steps, each with a precise purpose~\cite{seow2002writing}. This is a crucial process because it allows for analysis of ``what’s working, what isn’t, and what it still needs'' and adaptation to these needs along the way~\cite{jackson2022advantage}. In many cases, a needed change may only become apparent after much of the text is created, such as in the case of a reorganization or fixing inconsistencies or contradictions~\cite{vardi2012}. In this way, the current paradigm of generating text passages in a single pass can be severely limiting. 


\begin{figure*}
    \centering
    \includegraphics[width=\textwidth]{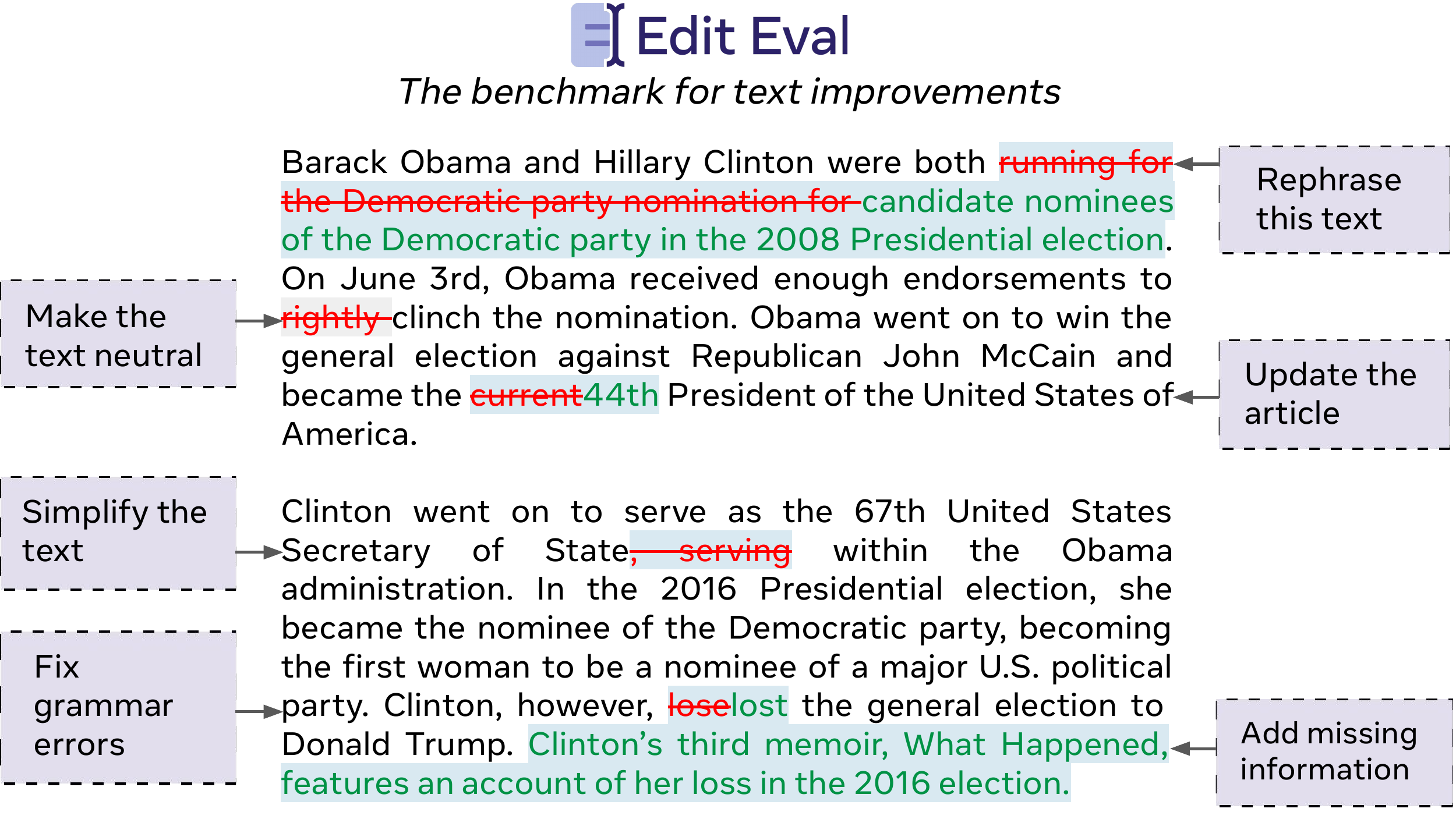}
    \caption{Example of neutralization, simplification, fluency, paraphrasing, and updating instructions and their corresponding expected edits. For illustrative purposes, we ground these examples in the same passage, but examples in \ee\ follow the format as described in Section~\ref{sec:instructions}.}
    \label{fig:demo}
\end{figure*}

Additionally, the current paradigm of continuous left-to-right generation is less controllable and not flexible to human-in-the-loop collaboration and feedback, and this absence of experienced human mediation in the writing process can be highly detrimental to the quality of the final product~\cite{greenberg2010editor}. While there are some existing production tools geared towards working with humans to compose articles and emails, such as Smart Compose from Google \footnote{https://www.blog.google/products/gmail/subject-write-emails-faster-smart-compose-gmail/} and text predictions from Microsoft \footnote{https://insider.office.com/en-us/blog/text-predictions-in-word-outlook}, these mostly focus on sentence completion and are not developed to improve upon prior text. A more powerful editing assistant, however, would not only be capable of providing recommendations for continuations of the text but also improvements upon the already existing text, such as making the tone more consistent, making diction more precise, or adding more engaging information. These AI tools should also permit iterative and non-sequential development of the text~\cite{seow2002writing}, which is naturally unavoidable, for example, if new or missing information or external references are required to update the text or if a reshuffling/rebalancing of text is needed.

In this work, we alternatively promote iterative text generation and improvement---successive iterations of modular additions and modifications of the text that are relevant to text editing such as making text clearer and adding missing information. Many datasets for natural language tasks are actually annotated at the sentence or paragraph level, rather than document or article level, naturally lending well to evaluating iterative edits. 


We create \ee, a benchmark and evaluation suite that leverages high-quality existing and new datasets for automatic evaluation of editing capabilities. Currently, many of these pertinent datasets live in separate packages and are often formatted in uniquely distinct ways. \ee\ downloads each dataset from their most recent version and standardizes each into a single format conducive to evaluation. Additionally, we include popular metrics for each task and a set of human-generated prompts to robustly measure a model's capability in executing the modular task when instructed. Figure~\ref{fig:demo} shows examples of such prompts and an example of a corresponding edit that we might expect for the given text. Using these prompts, we evaluate and compare several state-of-the-art language models, such as GPT-3~\cite{brown2020language}, OPT~\cite{zhang2022opt}, and PEER~\cite{peerpaper2022}. In summary, our contributions are as follows:
\begin{enumerate}
    \item We identify a set of tasks and datasets relevant to iterative text improvement and provide a pipeline to download and process these datasets into a single format.
    \item We open-source a publicly available instruction-based benchmark for automatic evaluation according to metrics commonly used for each editing task.
    \item We introduce a new dataset, \textsc{WAFER-insert},  for evaluating a model's capability to update information, which is based on the WAFER dataset~\cite{petroni-etal-2022-improving}.
    \item We provide a comparison of various state-of-the-art baselines evaluated on \ee\ at the dataset and prompt level.
\end{enumerate}

\section{Related Work}

Several multitask evaluation benchmarks have been open-sourced to the community to support progress in natural language understanding including GLUE~\cite{wang2018glue}, SuperGLUE~\cite{wang2019superglue}, decaNLP~\cite{mccann2018natural}, and GEM~\cite{gehrmann-etal-2021-gem}. These datasets, however, focus on a broad set of tasks in NLP (e.g., question answering, reading comprehension, and natural language inference). While all of these tasks are critical to natural language understanding, \ee\ focuses on curating a benchmark for measuring a model's capability to improve and edit text. 

There are several datasets which focus on iterative text revisions in the domain of Wikipedia~\cite{yang2017identifying, anthonio2020wikihowtoimprove}, academic essays~\cite{zhang2017corpus}, and news articles~\cite{spangher2022newsedits}. These works, however, focus on one particular domain and in some cases, a particular style like argumentative writing~\cite{zhang2017corpus}. \ee, on the other hand, includes examples from multiple domains: Wikipedia, Wikinews, news articles, and arXiv. \textsc{IteraTeR}~\cite{du2022understanding} is perhaps closet to \ee\ in that it provides iterative tasks from multiple domains, but it has a limited number of such tasks: fluency, coherence, clarity, style, and meaning-changed. Because this is a great starting point, we have included \textsc{IteraTeR} in \ee, and we additionally develop prompts for these tasks since \textsc{IteraTeR} is not instruction-based. Additionally, unlike \textsc{IteraTeR}, \ee\ includes novel datasets for tasks such as updating text using new information and neutralizing the text, which are core components of editing a factually-correct and unbiased article.

\section{The \ee\ Benchmark}

\ee\ is an instruction-based benchmark for iterative text generation/modification. \ee\ sources existing high-quality datasets---most with human annotations---containing tasks relevant to editing. These datasets are combined into a unified evaluation tool and can be evaluated with any metric provided in \ee. A task here refers to a type of edit (e.g., simplification or neutralization), and the specific task dictates which set of prompts to be used (e.g., simplify this text).

We consider seven editing tasks in \ee. The corresponding datasets for each task included in \ee\ are enumerated in Table~\ref{tab:data_overview}, along with the size of the test set. For ease of evaluation, we define a consistent format for all datasets in the \ee\ benchmark. Each dataset of every task has five core fields: ID, input text, gold edits, task type, and reference documents. The input text is the original text before revision, and the gold edits are the target edits for that specific task type. Lastly, the reference documents provide textual information from external articles or documents that are relevant to the task. The task that requires reference documents is updating, and otherwise, the reference documents field is empty.

The datasets in \ee\ were selected if they test a capability relevant to the art of editing and contain human-annotated gold edits, if possible. We also endeavored to include datasets that are broadly used by the community. The datasets in \ee\ are by no means exhaustive, but the \ee\ framework is flexible such that it can easily extend to new datasets and metrics in future versions.


\begin{table}
    \small
  \caption{Tasks, datasets, abbreviations used, and corresponding test size in \ee. The task type dictates which set of instructions are used. These are enumerated in Section~\ref{sec:prompts}.}
  \label{tab:data_overview}
  \begin{tabular}{l l l l l}
    \toprule
    Prompt & Dataset & Abbrev. & Size \\
    \midrule
    Clarity & \textsc{IteraTeR} & ITR-L & 185 \\
    Coherence & \textsc{IteraTeR} & ITR-O & 35 \\
    Fluency & \textsc{IteraTeR} & ITR-F & 88 \\
    Fluency & JFLEG & JFL & 747 \\
    Simplification & ASSET & AST &359\\
    Simplification & TurkCorpus & TRK & 359\\
    Paraphrasing & STS Benchmark & STS & 97 \\
    Neutralization & WNC & WNC & 1000 \\ 
    Updating & FRUIT & FRU & 914\\
    Updating & \textsc{WAFER-insert} & WFI & 4565 \\
  \bottomrule
     \end{tabular}
\end{table}


\subsection{Fluency, Clarity, and Coherence}

In this section, we describe the two datasets that compose this set of tasks: fluency (fixing grammatical or spelling errors), clarity (making the text clearer), and coherence (making the text more cohesive).

\paragraph{JFLEG} JHU FLuency-Extended GUG~\cite{napoles2017jfleg} focuses solely on the first task of fluency. JFLEG is based on the GUG (Grammatical vs Un-Grammatical) dataset~\cite{heilman2014predicting}, which is a dataset of sentences originally annotated for how grammatical the sentence is on a scale of 1 to 4. JFLEG builds upon the ungrammatical sentences in GUG and annotates each sentence with four corresponding corrected versions. 

\paragraph{\textsc{IteraTeR}} This dataset introduced by \citet{du2022understanding} contains both automatically-mined and human-annotated edits at the sentence and document-level. For our benchmark, we only utilize the sentence-level examples with human annotations. Additionally, \textsc{IteraTeR} has labels for the intent---the type of edit that produces the targets, which can be one of six classes: Fluency, coherence, clarity, style (conveying the writer's writing preferences), meaning-changed (updating or adding new information), and other (none of the others). We included all classes except style, meaning-changed, and other. We excluded style and other because these tasks had roughly 100 or less test examples, and the definitions were comparatively under-specified. We excluded meaning-changed because the task does not use reference documents for updating. This dataset is the only one in \ee\ that encompasses multiple tasks, and we refer to each respective subset using the abbreviations ITR-F (fluency), ITR-L (clarity), and ITR-O (coherence).

\subsection{Paraphrasing}


\paragraph{STSB} For paraphrasing, we use the STS benchmark from SemEval-2018~\cite{cer2017semeval}, which comprises English datasets used in the STS tasks of SemEval between 2012 and 2017. The selection of datasets includes text from image captions, news headlines and user forums. Each example contains an original sentence, a target sentence, and a similarity score indicating whether the target is a paraphrase of the original. This dataset is used for classification or regression, but for EditEval, we utilize all instances that we are confident are paraphrases, i.e., have the max similarity score of 5, as targets for generation evaluation. While other datasets such as ParaSCI~\cite{dong2021parasci} exist for paraphrase generation, these are automatically curated rather than human annotated, and \ee\ strives to utilize human-annotated datasets where possible.

\subsection{Simplification}


Simplification can be considered a very similar task to paraphrasing with the additional constraint that the output must be simpler than the input. The datasets we utilize for simplification are TurkCorpus~\cite{Xu-EtAl:2016:TACL} and ASSET~\cite{alva2020asset}.

\paragraph{TurkCorpus} This dataset, like ASSET, builds upon the Parallel Wikipedia Simplification (PWKP)~\cite{zhu-etal-2010-monolingual}. The PWKP dataset uses the Simple English Wikipedia and Standard English Wikipedia in parallel to create original-simplification pairs automatically. However, several works found PWKP to have a large proportion of targets that are not simplified or only partially aligned with the input~\cite{xu2015problems, amancio2014analysis, hwang2015aligning, vstajner2015deeper}, leading to the creation of a human-annotated corpus, TurkCorpus. TurkCorpus was manually created with eight reference simplifications for each original sentence in PWKP, but only used simplifications that are possible without deleting content or splitting sentences. 
\paragraph{ASSET} Because TurkCorpus encompassed only specific kinds of simplifications, this led to the creation of ASSET, which provides manually-produced simplifications through a much broader set of transformations. We include both in \ee, for the sake of comprehensiveness.

\subsection{Neutralization}

The task of neutralization refers to making a text more neutral. For example, in the sentence ``Obama was an excellent president who served two terms from 2008 to 2016'' the term \textit{excellent} violates Wikipedia's neutral point of view (POV) policy\footnote{https://en.wikipedia.org/wiki/Wikipedia:Neutral point of view}. For information-intensive content like Wikipedia and news articles in particular, reducing bias is crucial because bias is the single largest source of distrust in the media~\cite{jones2019online}. 
\paragraph{WNC} We use the Wiki Neutrality Corpus~\cite{pryzant2020automatically}, a collection of original and de-biased sentence pairs mined from Wikipedia edits by carefully filtering based on the editor's comments. While ideally we would like to include a human-annotated dataset, to our knowledge there does not exist a dataset for de-biasing article content at the sentence level.

\subsection{Updating}

In this section we describe the task of updating information which requires \textit{references}, text from external sources that are relevant to the particular task. Because of token-length restrictions, each external article is chunked into texts of fixed length. We limit the scope of the task to three chunks, and we refer to these selected chunks as our \textit{reference documents}. These references documents are represented in the edits by their index in the reference documents field (e.g., the first would be demarcated as [0]), and we discuss below how these reference documents were selected. 

\paragraph{\textsc{WAFER-insert}} The first dataset for updating information that we use is the WAFER dataset~\cite{petroni-etal-2022-improving}, which is a dataset collected from Wikipedia inline citations. Each instance of the original WAFER dataset contains a claim, the text surrounding the claim, and a set of external references, where the task is to choose one of the references to be cited after the claim. 
While the original intention of WAFER was to measure a system's capability to choose the correct citation, \ee\ utilizes WAFER for the task of inserting new information using content from the reference documents. We create \textsc{WAFER-insert}, which differs from WAFER in that the claim is deleted from the input. The goal here is to derive the original claim from the references and insert it into the text. For the reference documents, we select the top three chunks from the inline citation chunks that have the highest scores, using results from the verification engine introduced in~\cite{petroni-etal-2022-improving}.

\paragraph{FRUIT} In addition to \textsc{WAFER-insert}, we include the FRUIT dataset \citep{logan2021fruit}, a dataset collected by comparing two snapshots of a Wikipedia article where one contains updated or new information. The reference documents were identified by searching for other Wikipedia articles that provide evidence that supports the update. However, because there is no certainty that the identified evidentiary Wikipedia articles support the claim, the authors of FRUIT created a gold set by employing human annotation to filter out any new claims that are unsupported. We include this gold set in \ee, and only include reference documents if they actually appear in the output. Unlike \textsc{WAFER-insert}, the target edit contains not only the updated information but also the citation. For \ee, this is for verification purposes only, and the citation is removed when computing the metrics.

\section{Metrics}

The metrics we included in \ee\ are ones that are (1) shown to have significant correlation with human judgement for a task in \ee\ and (2) commonly used to benchmark one of the datasets in \ee. Below, we discuss each set of metrics in detail.

\paragraph{EM and EM-Diff}

Exact match (EM) is the percentage of examples for which the performed edit exactly matches any of the targets. EM-Diff is a variant of EM that is computed on the diff level, where diffs are obtained using Python's \texttt{difflib} library. For a model output $O$, we compute EM-Diff as follows:
$$\frac{|\text{diff}(I, R) \cap \text{diff}(I, O) |} {\max(|\text{diff}(I, R)|, |\text{diff}(I, O)|)}$$

\paragraph{SARI}

Introduced by \citet{Xu-EtAl:2016:TACL}, SARI is an n-gram based metric commonly used for measuring simplification~\cite{nisioi2017exploring, zhao2018integrating} and other editing tasks such as sentence fusion~\cite{malmi2019encode}. It has been demonstrated to correlate most closely with human judgement for simplification compared to many other n-gram based metrics~\cite{Xu-EtAl:2016:TACL}. The metric measures how simplified a candidate system output is relative to the original and to the simplification references by rewarding words added, kept, or deleted in both the target and the output. More specifically, this is done by computing the arithmetic mean of n-gram F1-scores for each of the three operations. We utilize the EASSE~\cite{alva2019easse} implementation of SARI, which addresses inconsistencies in the original implementation \footnote{https://github.com/feralvam/easse\#differences-with-original-sari-implementation}.

\paragraph{BLEU and iBLEU}

BLEU~\cite{papineni2002bleu} is another n-gram based metric that encourages a high proportion of n-gram matches between the output and the targets. BLEU, originally intended for machine translation, is very commonly used for many editing tasks such as simplification~\cite{vstajner2015deeper, sulem2018semantic} and improving fluency~\cite{pryzant2020automatically, du2022understanding}, and is shown to correlate well with human judgement of tasks such as grammaticality and meaning preservation~\cite{Xu-EtAl:2016:TACL}. 

For some tasks like simplification and paraphrasing, however, we require not only that the output is similar to the target, but that the output is sufficiently different from the input. iBLEU, a metric introduced by \citet{sun2012joint} is a weighted average of the BLEU score computed between the output and the targets and the negated BLEU score computed between the output and the input. More specifically, for a candidate output sentence O, human targets R, and an input text I, iBLEU is defined as:
$$ \alpha \times \text{BLEU}(O,R) - (1-\alpha) \times \text{BLEU}(O, I) $$
\citet{Xu-EtAl:2016:TACL} demonstrated that for simplification, iBLEU correlates on par or better with human judgement than BLEU does, though not as well as SARI on average.

\paragraph{GLEU}
GLEU~\cite{napoles2015ground} is another variant of BLEU frequently used for grammatical error correction~\cite{grundkiewicz2019neural, yuan2016grammatical, chollampatt2018multilayer}. The issue with using BLEU for minimal edits can be attributed to the difference between analyzing machine translation and editing tasks. In the former, an untranslated word should always be penalized, but in the editing setting, an unmodified word in both the target and the output does not necessarily need to be penalized. Unlike BLEU, GLEU is customized to penalize n-grams changed in the targets but left unchanged by the system output. \citet{napoles2015ground} not only demonstrated that GLEU correlates well with human rankings of corrections, but also that GLEU correlates much better than BLEU does. 

\paragraph{ROUGE and UpdateROUGE}

For the task of updating or adding new information, we follow \citet{logan2021fruit} and use ROUGE and UpdateROUGE~\cite{logan2021fruit}. ROUGE~\cite{lin2004rouge} is a popular n-gram based metric that is commonly used for evaluating summarization systems~\cite{ren2016redundancy, pasunuru2018multi}, but is also used in other tasks such as improving fluency~\cite{kann2018sentence} and simplification~\cite{vanderwende2006microsoft}. ROUGE essentially measures the overlap in n-grams between the system output and the targets. UpdateROUGE, a simple modification of ROUGE, computes ROUGE on the updated sentences rather than the full text. This is intended for tasks such as updating, because a majority of the target will remain unchanged. On the other hand, when evaluating using ROUGE, a system can often superficially achieve high scores by simply copying the input. 



\section{Baselines}

For each baseline, we use greedy decoding, and we do not perform any task-specific fine-tuning or in-context learning. We evaluate on \ee\ using the following baselines: 

\begin{itemize}
    \item \textbf{GPT-3} \citep{brown2020language} is a 175B parameter pretrained decoder-only model. We evaluate GPT-3 through OpenAI's API.\footnote{\url{https://beta.openai.com/}}
    \item \textbf{InstructGPT} \citep{ouyang2022training} is a  variant of GPT-3 that was fine-tuned on a large dataset of instructions and corresponding outputs written by humans. We evaluate the \emph{text-davinci-001} version described in \citep{ouyang2022training} since, at the time of writing, details about the training process for \emph{text-davinci-002} were not publicly available.
    \item \textbf{OPT} \citep{zhang2022opt} is an open-source replica of GPT-3. Like GPT-3, it is not fine-tuned on any labeled data.
    \item \textbf{T0}~\cite{sanh2022multitask} is a pretrained encoder-decoder model, which has demonstrated better performance than GPT-3 on several tasks despite being much smaller. It is initialized from the LM Adapt variant of T5~\cite{raffel2020exploring} and is fine-tuned on examples from 170 existing NLP datasets that are prompted using around 2000 crowdsourced prompts.
    \item \textbf{T0++} \citep{sanh2022multitask} is similar to T0, but trained on a few additional datasets from SuperGLUE~\cite{wang2019superglue}.
     \item \textbf{T$k$-Instruct} \citep{wang2022benchmarking} is similar to T0 and T0++ but instead fine-tuned on their dataset, Natural Instructions v2, a collection of instructions for more than 1,600 tasks, including grammatical error correction and text simplification.
     \item \textbf{PEER} \citep{peerpaper2022} A collaborative language model trained to infill parts of the writing process by leveraging self-training techniques. It is also initialized from the LM Adapt variant of T5, and further fine-tuned on edit histories from Wikipedia. We use the 3B and 11B PEER (SP) model (shortened here as PEER-3 and PEER-11, respectively), where SP refers to augmenting the training data with synthetic instructions and was shown to perform the best in \citet{peerpaper2022}.
\end{itemize}

\section{Instructions}
\label{sec:instructions}
\begin{figure}
    \centering
    \includegraphics[width=\linewidth]{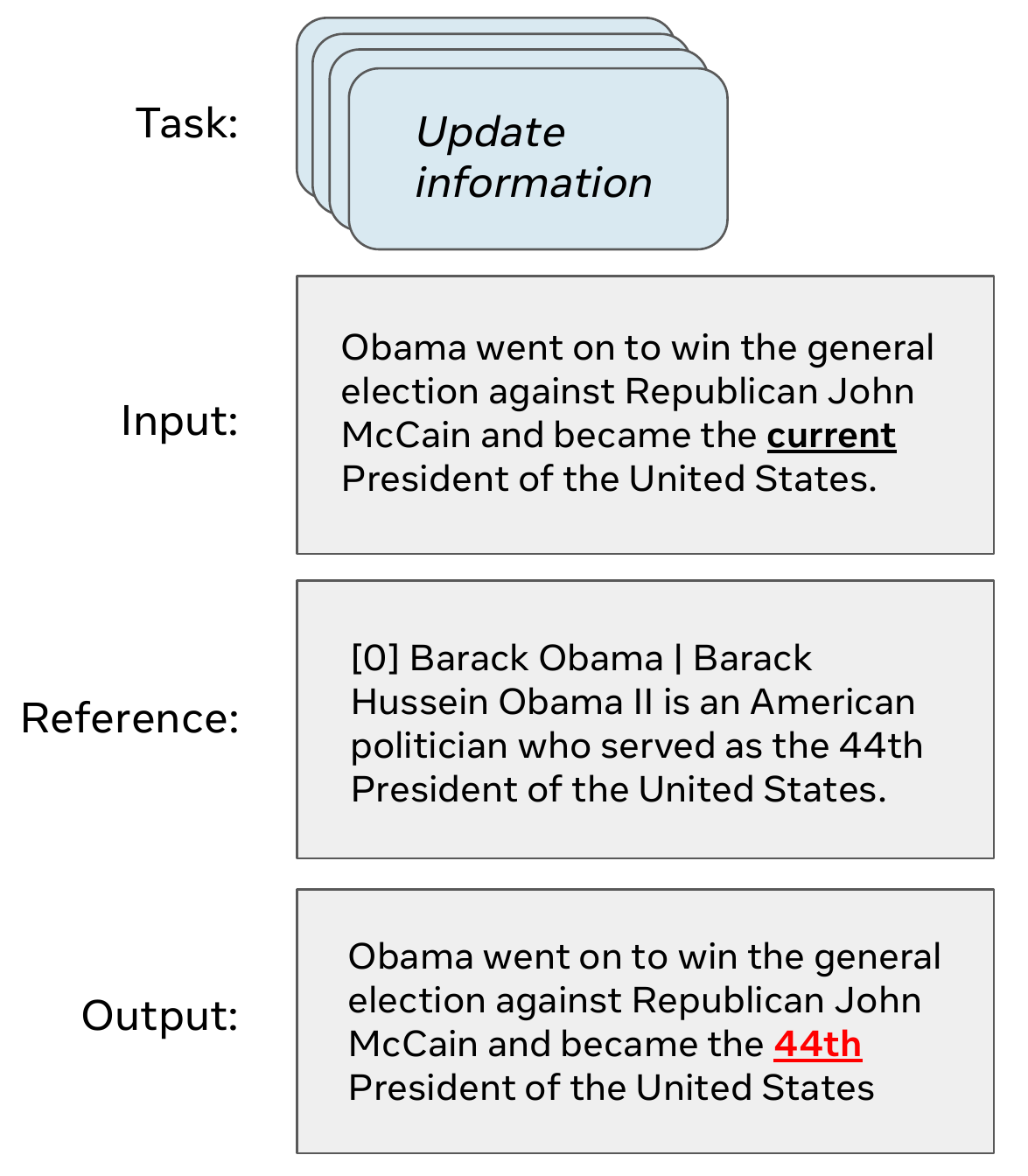}
    \caption{Example of inputs formatted when evaluating the baseline models. Each input is evaluated with a set of prompts that are determined by the task type.}
    \label{fig:input_format}
\end{figure}

We evaluate these baselines on their general capability to accomplish each task when prompted in natural language in a zero-shot fashion. Because there are a diverse set of ways in which to instruct for each task, we manually construct a set of 3--11 prompts in order to more robustly evaluate performance. For each task prompt $t$ and input $i$, the model is given a formatted input following the template: \begin{align*}
& \texttt{Task:}\ t \\
& \texttt{Input:}\ i \\
& \texttt{Output:}
\end{align*} with an additional field for references, should they be required. Figure~\ref{fig:input_format} shows an example of an input including references. For tasks without references, we exclude this field. Some slight modification to this template were made. For example, T$k$-Instruct expects the prompt to be prefixed by the string ``Definition:'' rather than ``Task:''). 

\section{Results}

\begin{figure*}[!ht]
    \centering
    \includegraphics[width=\linewidth]{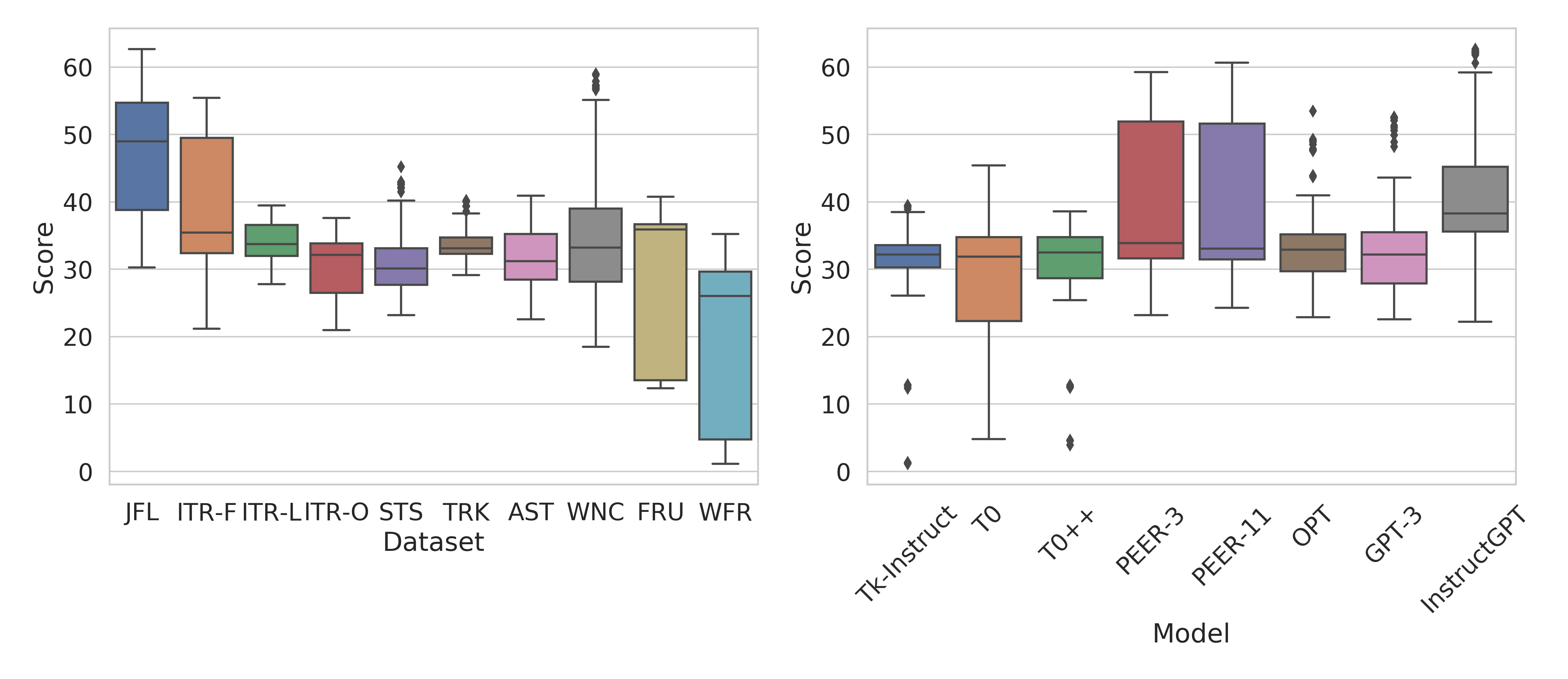}
    \caption{Left: Boxplot of SARI scores for each dataset averaged across models. Datasets which have the largest variance amongst the baselines are not necessarily harder tasks. Right: Boxplot of SARI scores for each baseline averaged across datasets. PEER has the largest range in performance across datasets, but OPT is the least robust to different prompts within a task (average coefficient of variation of 6.74\%, Table~\ref{tab:average_results}).}
    \label{fig:model_avg}
\end{figure*}

\begin{figure*}
     \centering
     \begin{subfigure}{\textwidth}
         \centering
         \includegraphics[width=\textwidth]{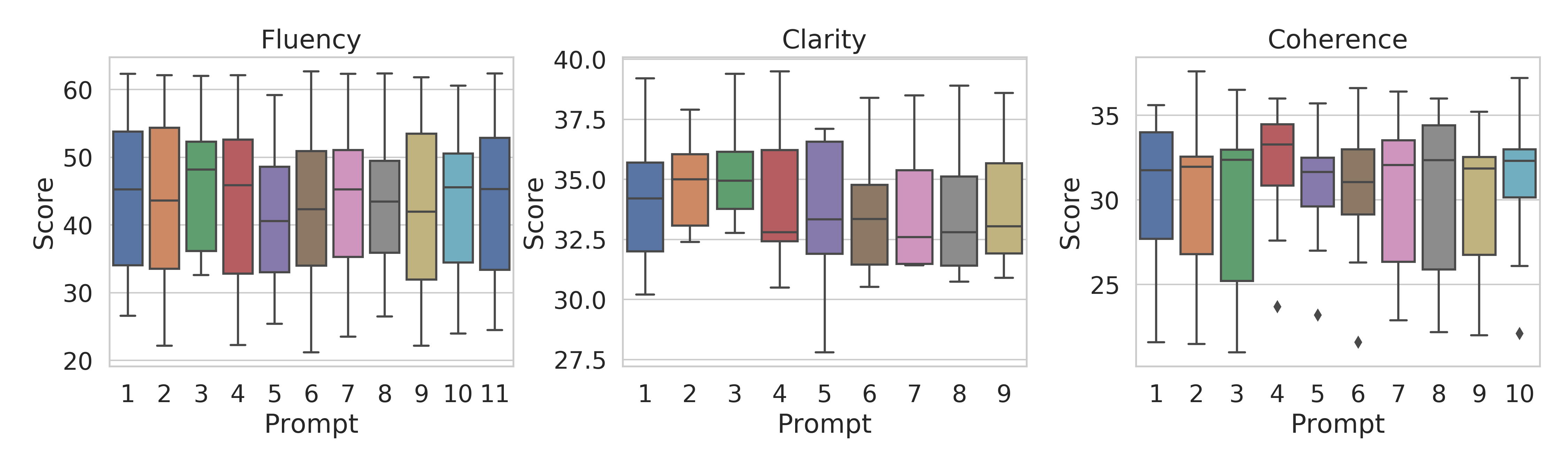}
     \end{subfigure}
     \hfill
     \vspace{-1.4\baselineskip}
     \begin{subfigure}{\textwidth}
         \centering
         \includegraphics[width=\textwidth]{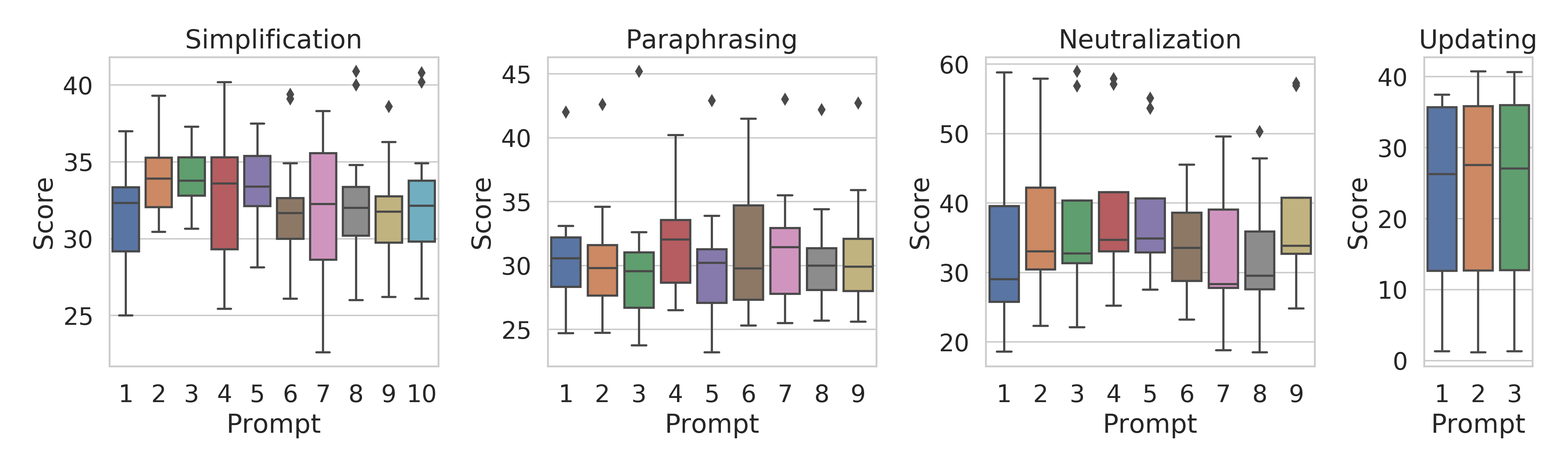}
     \end{subfigure}
     \hfill
     \vspace{-1.0\baselineskip}
     \caption{Boxplot of SARI scores for each prompt averaged across models. The prompts which achieve the maximum scores for each dataset (Table~\ref{tab:downstream_results_max_min}), are Prompts \#6 and 11 (fluency), 4 (clarity), 2 (coherence), 8 and 10 (simplification), 3 (paraphrasing), 2 (neutralization) and 2 and 1 (updating). These prompts do not necessarily exhibit high or low variance across models. Certain prompts evoke more variation across models due to factors such as using less frequently used language or being too unspecific.}
     \label{fig:prompt_avg}
\end{figure*}

We summarize results in Table~\ref{tab:average_results} with the aforementioned baselines averaged over all datasets and the breakdown for each dataset in Table~\ref{tab:downstream_results}. To visualize the variance according to each model, we show boxplots for each dataset and model according to the SARI metric in Figure~\ref{fig:model_avg}. Similarly, to visualize the variance for a given prompt, we present boxplots averaged across models in Figure~\ref{fig:prompt_avg}. We discuss several observations below.


\paragraph{InstructGPT and PEER perform the best overall.}  In Table~\ref{tab:average_results}, we show the mean SARI scores for each model averaged across all tasks using the average, maximum, and minimum scores across prompts. When using the average and minimum across prompts (third and fifth column, respectively) we see that InstructGPT performs the best overall, but when using the maximum score across prompts (fourth column), PEER-11 performs the best. Table~\ref{tab:downstream_results} enumerates the breakdown of the third column according to each dataset. In general, we see that InstructGPT achieves the highest scores with the exception of the updating and neutralization datasets, as well as ITR-F and ITR-L. For these datasets, the PEER models clearly outperform InstructGPT by a large margin, despite being nearly 60$\times$ smaller than InstructGPT and GPT-3. The substantially smaller models (T0, T0++, and T$k$-Instruct) struggle the most overall, even falling behind the copy baseline at times, except on ITR-L where T$k$-Instruct performs the best.

\begin{table}[ht]
\small
\centering
\begin{tabular}{llcccr} 
\toprule
 Model & Params & Avg. & Max & Min & CV\\
 \midrule
T$k$ & 3B & 28.2 & 30.1	& 26.1 & 4.65\\
T0 & 3B & 26.6  & 29.3 & 24.5 & 6.03\\
T0++ & 11B & 28.4 	& 30.3	& 26.7 & 5.13\\
PEER-3 & 3B & 38.8 	& 41.8	& 35.0 & 6.36\\
PEER-11 & 11B & 39.1 	& \textbf{42.1}	& 35.6 & 5.75\\
OPT & 175B & 32.8	& 36.4	& 29.0 & 6.70\\
GPT-3 & 175B & 32.8 & 35.8	& 29.4  & 6.74\\
InstructGPT & 175B & \textbf{39.6} & 41.3 & \textbf{37.4} & \textbf{3.60}\\
  \bottomrule
\end{tabular}
\caption{Mean SARI scores across all tasks using the average across prompts (Avg.), the maximum across prompts (Max), and the minimum across prompts (Min). The coefficient of variance (CV), computed as the standard deviation across prompts normalized by the average, is shown in the final column. Best values are in bold. When using averages across prompts and using the minimum, InstructGPT performs the best, but PEER performs the best when using the maximum across prompts.}
\label{tab:average_results}
\end{table}

\begin{table*}[ht]
\newcolumntype{Y}{>{\centering\arraybackslash}X}
    \newcommand{\negphantom}[1]{\settowidth{\dimen0}{#1}\hspace*{-\dimen0}}
    \small
    \setlength{\tabcolsep}{4.5pt}
    \begin{tabularx}{\linewidth}{lccccccccccccc}
    \toprule
    & \multicolumn{2}{c@{\hskip 0.1cm}}{\textit{Fluency}} &  \multicolumn{1}{c@{\hskip 0.1cm}}{\textit{Clarity}} & \multicolumn{1}{c@{\hskip 0.1cm}}{\textit{Coherence}} & \multicolumn{1}{c@{\hskip 0.1cm}}{\textit{Para.}} &  \multicolumn{2}{c@{\hskip 0.1cm}}{\textit{Simplification}}  & \multicolumn{1}{c@{\hskip 0.1cm}}{\textit{Neutral.}}& \multicolumn{2}{c@{\hskip 0.1cm}}{\textit{Updating}} & \\
    \textbf{Model} & JFL & ITR-F & ITR-L & ITR-O & STS & TRK & AST & WNC & FRU & WFI \\
    \midrule
    Copy & 26.7 / 40.5 & 32.3 / 86.0 & 29.5 / 62.9 & 31.3 / 77.2 & 21.1 & 26.3 & 20.7 & 31.9 / \phantom{0}0.0 & 29.8 / \phantom{0}0.0 & 33.6 / --\negphantom{--}\phantom{0.0}\\ 
    \midrule
T$k$ & 31.8 / 39.0 & 32.4 / 61.6 & \textbf{38.4} / \textbf{58.4} & 33.8 / \textbf{70.4} & 30.2 & 32.8 & 29.9 & 31.3 / \phantom{0}0.4 & 12.6 / \phantom{0}3.6 & \phantom{0}1.3 / \phantom{0}4.5\\ 
T0 & 42.0 / 38.8 & 24.6 / 34.9 & 32.6 / 30.2 & 22.2 / 21.6 & 34.3 & 34.4 & 32.3 & 22.3 / \phantom{0}0.0 & 14.2 / \phantom{0}9.6 & \phantom{0}5.1 / 16.3 \\ 
T0++ & 34.7 / 43.2 & 35.3 / 75.8 & 37.6 / 56.5 & 32.7 / 59.9 & 28.4 & 32.9 & 28.2 & 29.3 / \phantom{0}0.3 & 12.6 / \phantom{0}3.7 & \phantom{0}4.4 / \phantom{0}8.1 \\ 
PEER-3 & 55.5 / 54.3 & 51.4 / 84.3 & 32.1 / 47.1 & 32.1 / 59.8 & 28.6 & 32.5 & 30.5 & 53.3 / 21.6  & 39.1 / 30.9 & 34.4 / 18.7 \\ 
PEER-11 & 55.8 / 54.3 & \textbf{52.1} / \textbf{85.2} & 32.5 / 51.3 & 32.7 / 62.7 & 28.2 & 32.1 & 29.5 & \textbf{54.5} / \textbf{22.8} & \textbf{39.6} / \textbf{31.4} & \textbf{34.9} / \textbf{20.4}\\ 
OPT & 47.3 / 47.5 & 34.7 / 70.6 & 31.5 / 31.5 & 27.6 / 36.1 & 29.1 & 32.6 & 31.8 & 31.2 / \phantom{0}0.4 & 35.9 / 27.3 & 26.7 / 11.2 \\ 
GPT-3 & 50.3 / 51.8 & 32.1 / 56.7 & 33.5 / 39.7 & 26.9 / 36.1 & 27.2 & 33.0 & 30.5 & 31.7 / \phantom{0}0.6 & 36.0 / 21.5 & 27.2 / 10.6 \\
InsGPT & \textbf{61.8} / \textbf{59.3} & 48.8 / 82.7 & 35.1 / 48.4 & \textbf{35.9} / 60.2 & \textbf{42.5} & \textbf{38.8} & \textbf{38.0} & 35.4 / \phantom{0}2.2 & 36.3 / 24.7 & 23.6 / 16.1 \\  
\midrule
SotA & \negphantom{--}\phantom{00.0}-- / \textit{62.4} &  \textit{37.2} / --\negphantom{--}\phantom{00.0} &  \textit{46.2} / --\negphantom{--}\phantom{00.0} &  \textit{38.3} / --\negphantom{--}\phantom{00.0} & \multicolumn{1}{c}{--} & \textit{34.4} &  \textit{37.2}  & \negphantom{--}\phantom{00.0}-- / \textit{45.8} & \negphantom{--}\phantom{00.0}-- / \textit{47.4} & \negphantom{--}\phantom{00.0}-- / --\negphantom{--}\phantom{00.0}  \\ 
    \bottomrule
    \end{tabularx}
    \caption{Results for all datasets, averaged across prompts. T$k$-Instruct and InstructGPT are shorthanded as T$k$ and InsGPT, respectively. The first numbers for each task are SARI scores; additional metrics are GLEU for fluency, clarity, and coherence, EM for neutralization, Update-R1 for updating. Supervised scores from left to right are from \citet{ge2018reaching}, \citet{du2022understanding},
    \citet{martin2020muss}, \citet{pryzant2020automatically} and \citet{logan2021fruit}, respectively. The best result for each dataset is shown in bold.}
    \label{tab:downstream_results}
\end{table*}

\paragraph{Most baselines lag substantially behind the supervised SOTA, especially in the task of updating and neutralization.} 
We show the supervised state-of-the-art results in the final row of Table~\ref{tab:downstream_results}, which in almost all cases surpasses the performance of the best baseline. The gap is largest for the tasks of neutralization and updating (34--50\% decrease from the supervised SOTA to the best baseline scores), whereas for other tasks, this decrease is only within 5--14\%. It is conceivable that the difficulty with these two particular tasks is a consequence of the comparatively fewer datasets and research devoted to them compared to that of the more mainstream NLP tasks, such as paraphrasing. 


\paragraph{Tasks that are the most challenging are not necessarily ones with the highest variance across models.} 
In observing Figure~\ref{fig:model_avg} (left), we see that the tasks which have the largest variance across models (assessed using the interquartile range (IQR)) are fluency and updating. This is despite the fact that the fluency datasets are arguably easier (i.e., many of the models come close to the supervised SOTA) than the updating datasets, exemplifying that difficulty and robustness can be independent axes. JFLEG also appears to be easier than ITR-F (average SARI score of 45.1 versus 38.2). This is not surprising since JFLEG sources from the TOEFL exam, which has primarily simpler and conversational sentences, whereas \textsc{IteraTeR} is composed of technical sentences from Wikipedia, ArXiv, and Wikinews. Likewise, TurkCorpus seems on average to be slightly easier than ASSET, which is expected since it was created to be more diverse than TurkCorpus. 

\paragraph{PEER has the highest variance across all tasks, but OPT and GPT-3 are the least robust to different prompts.} 

From Figure~\ref{fig:model_avg} (right), we observe that the PEER models have the largest range in performance from dataset to dataset. Within each task, however, GPT-3 and OPT have the highest coefficient of variation or standard deviation normalized by the mean (6.74\% and 6.70\%, respectively), whereas for the 3B and 11B PEER models, these values are smaller (6.36\% and 5.75\%), as enumerated in Table~\ref{tab:average_results}. This could be a consequence of the fact that GPT-3 and OPT are not trained explicitly to follow instructions, whereas the remaining baselines are. 


\begin{figure}
    \centering
    \includegraphics[width=\linewidth]{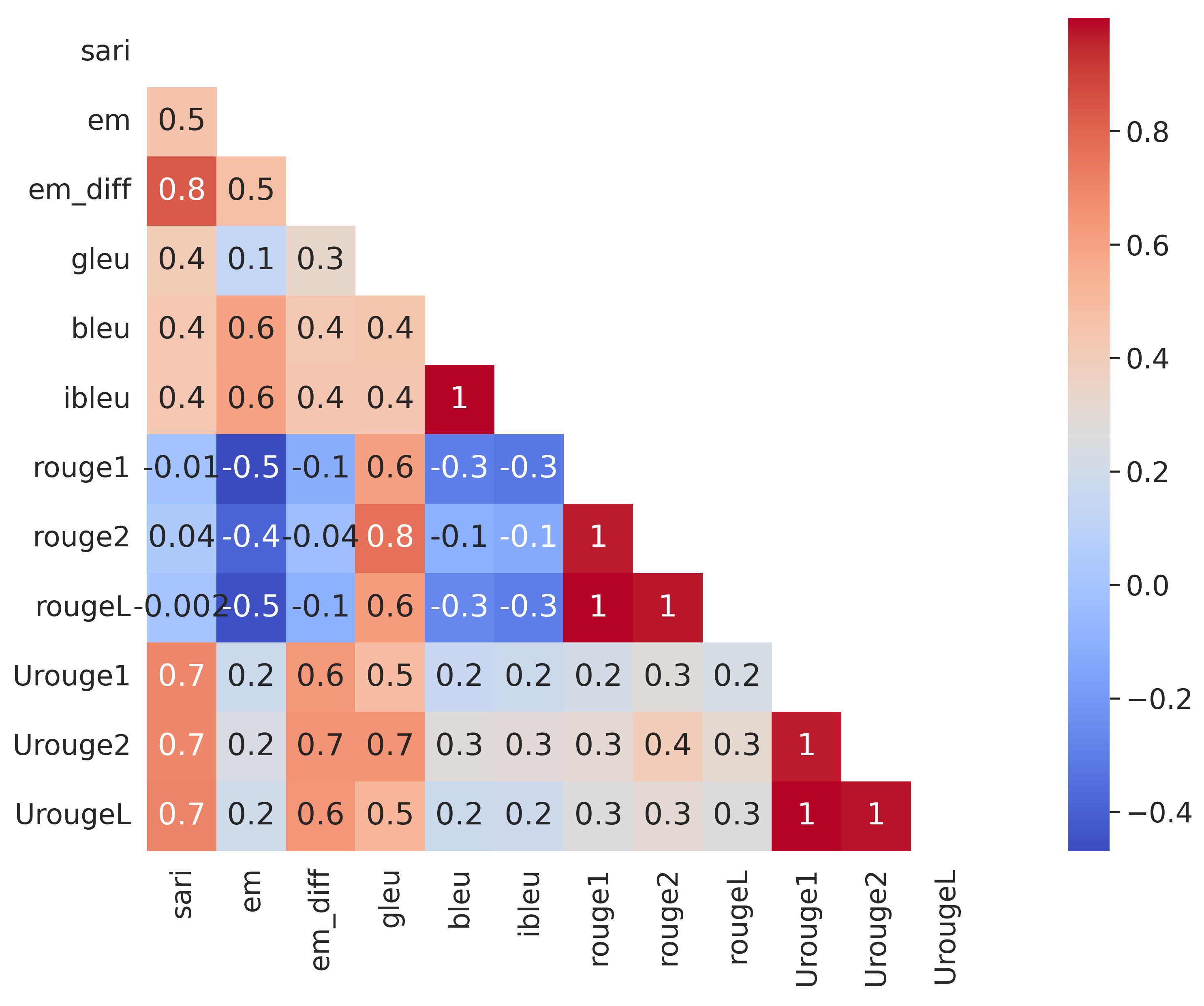}
    \caption{Pearson correlation between metrics using data for all datasets except WAFER and FRUIT and all baselines except PEER. Different families of metrics can have low correlation and even conflict, at times.}
    \label{fig:metric_corr}
\end{figure}

\paragraph{Prompts chosen according to maximum performance and prompts chosen according to robustness across models can be different.} Ideally, we would like to create prompts that are not only robust to different models, but achieve the highest performance using the best baseline. In assessing variance from Figure~\ref{fig:prompt_avg}, we see that certain prompts stand out as less robust relative to others. For example, for neutralization, Prompts \#1, 2, and 7 are less robust likely because they use uncommonly used language such as ``Remove points of views'' or ``Neutralize this text''. Some of the prompts which are less robust for simplification (Prompts \#4, 7) and paraphrasing (Prompts \#4, 6) are sometimes ones with less specific commands such as ``Rewrite this text'' versus ``Rewrite this with different wording''---in the case of the former, an empirical assessment shows that the models seem to more often copy the original text and make fewer modifications. Unfortunately, choosing prompts that are the most robust, does not always entail prompts which achieve the maximum score---Prompt \#5 for clarity achieves the maximum but has the largest IQR. Some of the tasks exhibit a great degree of outlier behavior (coherence, paraphrasing, or neutralization), which is either due to T0 performing exceedingly low or InstructGPT/PEER performing exceedingly well. Other tasks such as fluency and updating seem to have prompts with roughly a similar range of performance.

\paragraph{Different metrics do not always correlate well with each other.} We measure the Pearson correlation between each pair of metrics using evaluation scores for all baselines, which is shown in Figure~\ref{fig:metric_corr} as a heatmap. We exclude PEER in this analysis since it shows exceedingly strong performance in some cases, and we exclude the updating datasets since they are of a very different nature from the other datasets. We find that while families of variants like BLEU and iBLEU as well as ROUGE and UpdateROUGE show strong correlation within each set ($>$ 0.97), the two sets are inversely correlated with one another (-0.29 to -0.1). ROUGE actually appears to be the metric that most conflicts with all other metrics, whereas GLEU seems to be the metric that is most in harmony with the rest (0.41--0.76). Though SARI is not correlated with ROUGE, it is the metric which shows the strongest correlation with EM-Diff (0.83) and UpdateROUGE (0.7).


\section{Discussion}

We present \ee, a benchmark composed of handcrafted, task-specific instructions for several editing datasets across multiple domains. \ee\ is a means of evaluating models for these tasks according to multiple popular metrics, all within a single, unified tool. We show that while state-of-the-art models such as InstructGPT and PEER have impressive performance, in general the baselines lag behind the supervised state-of-the-art, particularly for the task of updating and neutralization. Our analysis of metrics and prompts shows that several popular metrics are not well-correlated, even conflicting at times, and that small changes in the wording of a prompt can lead to substantial changes in performance and robustness across models. This suggests further work is needed to develop models comprehensively capable of executing editing tasks in addition to developing a standardized way of measuring editing capabilities and systematically selecting prompts. In releasing this work, we hope to bolster work in which language models are utilized for text generation that is iterative, and therefore potentially more controllable, collaborative, and capable of revising and correcting text. 

\paragraph{Limitations}

Our evaluation tool is by no means an exhaustive measurement of editing capabilities. Firstly, there are additional domains that could potentially be added to \ee, such as books and blogs; as it currently stands, \ee\ is primarily constructed from the domain of Wikipedia. Fortunately, \ee's framework is flexible to the addition of datasets, provided that it has an input and target edit. In the same spirit, there are additional editing tasks such as verifying facts, citing, and reorganizing sentences/paragraphs which would be valuable to include in \ee. While we recognize these tasks as valuable to include in \ee, we consider these to be out of scope for the work at hand. Finally, our results demonstrate that many of the metrics give conflicting signal as to the rankings of the baselines, indicating further work is needed to identify better metrics for measuring overall editing capacity. 

\bibliography{anthology,editeval}
\bibliographystyle{acl_natbib}

\clearpage
\appendix

\section{Domains}

In \ee\ we strive to encompass datasets from many different domains, with an emphasis on factual content. Below in Table~\ref{tab:data_domains}, we enumerate these domains.

\begin{table}[htb]
  \caption{Number of targets provided ($|T|$) and the domains covered by each dataset.}
  \small
  \label{tab:data_domains}
  \begin{tabular}{l l l l l}
    \toprule
    Dataset & $|T|$  & Domains\\
    \midrule
    \textsc{IteraTeR} & 1 & Wikipedia, ArXiv, and Wikinews\\
    JFLEG & 4 & TOEFL exam\\
    WNC  & 1 & Wikipedia\\ 
    STS Benchmark & 1 & Wikipedia, Q\&A, news forums,\\
    & & videos, image descriptions\\
    ASSET & 10 & Wikipedia\\
    TurkCorpus & 8 & Wikipedia\\
     \midrule
    \textsc{WAFER} & 1 & Wikipedia\\
    FRUIT & 1 & Wikipedia\\
  \bottomrule
     \end{tabular}
\end{table}

\section{Prompts}
\label{sec:prompts}
Below we enumerate the prompts used in \ee\ for each task. We also present Table~\ref{tab:downstream_results_max_min} which shows the max and min results across these prompts as opposed to the average in Table~\ref{tab:downstream_results}.

\paragraph{Fluency}
\begin{enumerate}[nolistsep]
\item Fix grammar errors
\item Fix grammar or spelling mistakes
\item Fix grammar errors in this sentence
\item Fix all grammatical errors
\item Fix errors in this text
\item Update to remove grammar errors
\item Remove all grammatical errors from this text
\item Improve the grammar of this text
\item Grammar improvements
\item Remove grammar mistakes
\item Fix the grammar mistakes
\end{enumerate}
\paragraph{Clarity}
\begin{enumerate}[nolistsep]
\item Make the text more formal, concise, readable and understandable
\item Make the text more formal
\item Make the text more concise
\item Make the text more readable
\item Improve the readability of the text
\item Make the text more understandable
\item Make the text clearer
\item Make the text easier to understand
\item Improve the clarity of the text
\end{enumerate}
\paragraph{Coherence}
\begin{enumerate}[nolistsep]
\item Make the text more cohesive, logically linked and consistent as a whole
\item Make the text more cohesive
\item Improve the cohesiveness of the text
\item Make the text more logical
\item Make the text more consistent
\item Improve the consistency of the text
\item Make the text more understandable
\item Make the text clearer
\item Make the text easier to understand
\item Improve the coherency of the text
\end{enumerate}
\paragraph{Neutralization}
\begin{enumerate}[nolistsep]
\item Remove POV
\item Neutralize this text
\item Make this more neutral
\item Make this text more neutral
\item Make this paragraph more neutral
\item Remove unsourced opinions from this text
\item Remove non-neutral points of view
\item Remove points of view
\item Make this text less biased
\end{enumerate}
\paragraph{Paraphrasing}
\begin{enumerate}[nolistsep]
\item Paraphrase this sentence
\item Paraphrase
\item Paraphrase this paragraph.
\item Use different wording
\item Paraphrase this text
\item Rewrite this text
\item Rewrite this text with different wording
\item Rephrase this text
\item Reword this text
\end{enumerate}

\paragraph{Simplification}
\begin{enumerate}[nolistsep]
\item Simplify this sentence
\item Make this simpler
\item Simplify
\item Make this easier to understand
\item Simplification
\item Change to simpler wording
\item Simplify this paragraph.
\item Use simpler wording
\item Simplify this text
\item Make this text less complex
\end{enumerate}

\paragraph{Updating}
\begin{enumerate}[nolistsep]
\item Add missing information
\item Update the article
\item Update with new information
\end{enumerate}

\begin{table*}[t!]
\newcolumntype{Y}{>{\centering\arraybackslash}X}
    \newcommand{\negphantom}[1]{\settowidth{\dimen0}{#1}\hspace*{-\dimen0}}
    \small
    \setlength{\tabcolsep}{4.5pt}
    \begin{tabularx}{\linewidth}{lccccccccccccc}
    \toprule
    & \multicolumn{2}{c@{\hskip 0.1cm}}{\textit{Fluency}} &  \multicolumn{1}{c@{\hskip 0.1cm}}{\textit{Clarity}} & \multicolumn{1}{c@{\hskip 0.1cm}}{\textit{Coherence}} & \multicolumn{1}{c@{\hskip 0.1cm}}{\textit{Para.}} & \multicolumn{2}{c@{\hskip 0.1cm}}{\textit{Simplification}}  & \multicolumn{1}{c@{\hskip 0.1cm}}{\textit{Neutral.}}& \multicolumn{2}{c@{\hskip 0.1cm}}{\textit{Updating}} & \\
    \textbf{Model} & JFL & ITR-F & ITR-L & ITR-O & STS & TRK & AST & WNC & FRU & WFI \\
    \midrule
T$k$ & 32.9 / 41.6 & 36.0 / 77.6 & \textbf{39.5} / \textbf{63.3} & 35.7 / \textbf{77.1} & 33.1 & 34.9 & 32.6 & 33.8 / \phantom{0}1.3 & 12.9 / \phantom{0}4.1 & \phantom{0}1.3 / \phantom{0}5.0 \\ 
T0 & 45.4 / 43.1 & 32.6 / 50.9 & 33.8 / 34.0 & 23.7 / 25.5 & 35.9 & 35.3 & 35.9 & 27.5 / \phantom{0}0.1 & 14.9 / 12.4 & \phantom{0}5.4 / 17.2 \\ 
T0++ & 36.7 / 43.9 & 37.2 / 82.0 & 38.6 / 61.6 & 36.0 / 75.8 & 30.7 & 33.9 & 33.3 & 32.1 / \phantom{0}0.6 & 12.8 / \phantom{0}3.7 & \phantom{0}4.6 / \phantom{0}8.5 \\ 
PEER-3 & 59.3 / 57.7 & 54.5 / 86.3 & 34.0 / 60.6 & 33.8 / 74.1 & 34.6 & 36.4 & 35.5 & 57.4 / 29.3 & 40.2 / \textbf{33.6} & 34.7 / 20.2 \\ 
PEER-11 & 60.6 / 59.4 & \textbf{55.4} / \textbf{87.0} & 34.4 / 61.4 & 34.5 / 75.8 & 33.1 & 35.7 & 33.9 & \textbf{59.0} / \textbf{30.9} & \textbf{40.8} / 33.4 & \textbf{35.2} / \textbf{21.4} \\ 
OPT & 53.5 / 53.9 & 41.0 / 78.5 & 35.6 / 44.4 & 34.4 / 56.9 & 31.1 & 34.7 & 35.3 & 34.9 / \phantom{0}0.9 & 35.9 / 28.1 & 27.0 / 12.3 \\ 
GPT-3 & 52.6 / 54.2 & 39.1 / 79.2 & 35.6 / 45.8 & 29.9 / 42.9 & 29.4 & 35.5 & 35.9 & 34.9 / \phantom{0}1.1 & 36.3 / 21.6 & 28.2 / 11.2 \\
InsGPT & \textbf{62.7} / \textbf{60.4} & 51.0 / 85.0 & 36.5 / 52.6 & \textbf{37.6} / 68.8 & \textbf{45.2} & \textbf{40.2} & \textbf{40.9} & 37.2 / \phantom{0}3.8 & 36.6 / 25.2 & 26.0 / 17.3 \\ 
\midrule
T$k$ & 30.3 / 35.9 & 27.9 / 42.1 & 36.8 / 49.9 & 32.2 / \textbf{63.4} & 28.6 & 30.6 & 26.1 & 27.9 / \phantom{0}0.0 & 12.3 / \phantom{0}3.4 & \phantom{0}1.2 / \phantom{0}4.1 \\ 
T0 & 39.5 / 34.2 & 21.2 / 26.7 & 31.4 / 27.4 & 21.0 / 18.0 & 31.9 & 32.9 & 27.6 & 18.5 / \phantom{0}0.0 & 13.7 / \phantom{0}8.1 & \phantom{0}4.8 / 15.6 \\ 
T0++ & 33.0 / 42.2 & 33.1 / 62.3 & \textbf{36.8} / \textbf{52.6} & 29.3 / 45.8 & 25.5 & 31.9 & 25.4 & 27.4 / \phantom{0}0.2 & 12.5 / \phantom{0}3.7 & \phantom{0}3.9 / \phantom{0}7.5 \\ 
PEER-3 & 50.2 / 49.8 & 45.4 / 77.2 & 30.5 / 36.7 & 31.1 / 47.3 & 23.2 & 29.1 & 25.4 & 44.4 / 13.5 & 37.0 / 26.5 & 34.1 / 16.3 \\ 
PEER-11 & 49.8 / 46.7 & \textbf{45.9} / \textbf{82.5} & 31.4 / 43.3 & 31.9 / 47.9 & 24.3 & 29.4 & 25.7 & \textbf{45.5} / \textbf{15.7} & \textbf{37.5} / \textbf{27.3} & \textbf{34.7} / \textbf{19.0}\\ 
OPT & 40.7 / 41.0 & 29.7 / 55.5 & 27.8 / 22.1 & 22.9 / 24.6 & 26.1 & 30.3 & 26.2 & 25.0 / \phantom{0}0.0 & 35.8 / 26.6 & 26.5 / \phantom{0}9.8 \\ 
GPT-3 & 43.6 / 46.7 & 27.8 / 41.3 & 32.2 / 35.8 & 24.4 / 28.8 & 25.3 & 29.3 & 22.6 & 26.0 / \phantom{0}0.2 & 35.6 / 21.2 & 26.1 / 10.0 \\
InsGPT & \textbf{59.2} / \textbf{56.2} & 44.7 / 77.4 & 34.1 / 44.3 & \textbf{33.4} / 53.0 & \textbf{40.2} & \textbf{37.0} & \textbf{35.4} & 32.4 / \phantom{0}0.7 & 35.9 / 24.4 & 22.2 / 15.3 \\ 
\midrule
Copy & 26.7 / 40.5 & 32.3 / 86.0 & 29.5 / 62.9 & 31.3 / 77.2 & 21.1 & 26.3 & 20.7 & 31.9 / \phantom{0}0.0 & 29.8 / \phantom{0}0.0 & 33.6 / --\negphantom{--}\phantom{0.0}\\ 
SotA & \negphantom{--}\phantom{00.0}-- / \textit{62.4} &  \textit{37.2} / --\negphantom{--}\phantom{00.0} &  \textit{46.2} / --\negphantom{--}\phantom{00.0} &  \textit{38.3} / --\negphantom{--}\phantom{00.0} & \multicolumn{1}{c}{--} & \textit{34.4} &  \textit{37.2}  & \negphantom{--}\phantom{00.0}-- / \textit{45.8} & \negphantom{--}\phantom{00.0}-- / \textit{47.4} & \negphantom{--}\phantom{00.0}-- / --\negphantom{--}\phantom{00.0}  \\ 
    \bottomrule
    \end{tabularx}
\caption{Maximum (top half) and minimum (bottom half) scores across prompts for all downstream tasks considered. The first numbers for each task are SARI scores; additional metrics are GLEU for fluency, clarity, and coherence, EM for neutralization, Update-R1 for updating. The best results are highlighted in bold. T$k$-Instruct and InstructGPT are shorthanded as T$k$ and InsGPT, respectively.}
    \label{tab:downstream_results_max_min}
\end{table*}

\end{document}